\documentclass[11pt,a4paper]{article}

\usepackage[utf8]{inputenc}
\usepackage[T1]{fontenc}

\usepackage{amsmath,amssymb}
\usepackage{graphicx}
\usepackage{booktabs}
\usepackage[margin=2.5cm]{geometry}
\PassOptionsToPackage{hyphens}{url}
\usepackage[colorlinks=true, linkcolor=blue, citecolor=blue, urlcolor=blue]{hyperref}
\usepackage{orcidlink}
\usepackage[numbers,sort&compress]{natbib}

\title{TRACE: A Metrologically-Grounded Engineering Framework\\
       for Trustworthy Agentic AI Systems in\\
       Operationally Critical Domains}

\author{%
  Serhii Zabolotnii\,\orcidlink{0000-0003-0242-2234}%
  \thanks{Corresponding author. Email:
    \href{mailto:zabolotnii.serhii@csbc.edu.ua}{zabolotnii.serhii@csbc.edu.ua}.}%
}

\date{%
  \normalsize
  Cherkasy State Business College, Cherkasy 18028, Ukraine\\
  State Scientific Research Institute of Armament and Military Equipment\\
  \quad Testing and Certification, Cherkasy, Ukraine\\
  healthPrecision, New York, NY, USA
}

\begin{document}

\maketitle

\begin{abstract}
We introduce TRACE, a cross-domain engineering framework for trustworthy agentic AI
in operationally critical domains.
TRACE combines a four-layer reference architecture with an explicit classical-ML
vs.\ LLM-validator split (L2a/L2b), a stateful orchestration-and-escalation policy (L3),
and bounded human supervision (L4);
a metrologically grounded trust-metric suite mapped to GUM/VIM/ISO~17025;
and a Model-Parsimony principle quantified by the Computational Parsimony Ratio (CPR).
Three instantiations---clinical decision support, industrial multi-domain
operations, and a judicial AI assistant---transfer the same architecture and metrics
across principally different governance contexts.
The L2a/L2b separation makes the use of large language models a deliberate design
decision rather than an architectural default, with parsimony quantified through CPR.
TRACE introduces CPR as a first-class design principle in trustworthy-AI engineering.
\end{abstract}


\medskip
\noindent\textbf{Companion site:}
Documentation, the up-to-date publication roster, and CRediT-style author
contributions are available at \url{https://traces.solutions}.

% ─────────────────────────────────────────────────────────────────────────────
\section{Introduction}
\label{sec:intro}
% ─────────────────────────────────────────────────────────────────────────────

Trust in agentic AI is not an intrinsic property of the model.
A high-benchmark model can behave dangerously in real-world deployment;
a modestly accurate model can operate safely within a correctly engineered architecture.
As agentic AI systems enter operationally critical domains---clinical decision support,
industrial multi-domain operations, judicial assistance, energy and infrastructure
management---this observation becomes an engineering problem rather than a philosophical one:
\emph{how should the architecture be designed so that trust is a measurable property of
the system, not merely a declaration?}

The 2024--2026 literature addresses this question through four parallel but largely
disconnected streams.

\textbf{Stream~1 -- Clinical trust frameworks.}
FUTURE-AI~\cite{lekadir2025futureai} provides an international six-principle consensus
for trustworthy clinical AI
(fairness, universality, traceability, usability, robustness, explainability).
TRIAD~\cite{li2026triad} formulates a three-component governance/value/deployment model
for human--AI collaboration in medicine.
Both are domain-specific and lack metrological grounding or cross-domain applicability.

\textbf{Stream~2 -- Hierarchical agentic oversight.}
Tiered Agentic Oversight (TAO) by Kim et al.~\cite{kim2025tao} formalises tiered agentic
escalation with an orchestrator and clinician in the loop, empirically demonstrating
absorption of 16.9--24.3\% of individual-agent errors (Table~6, \cite{kim2025tao}).
TAO is the closest architectural predecessor of TRACE and an empirical anchor for
cross-domain generalisation.
A complementary, more recent line of work---Eywa by Li et al.~\cite{li2026eywa}---
demonstrates heterogeneous foundation-model / LLM collaboration through an explicit
FM--LLM communication protocol (\emph{Tsaheylu}), with the EywaBench benchmark
showing consistent token-cost reduction at improved utility across nine scientific
domains.
However, TAO (i)~is limited to the clinical domain; (ii)~implements routing and rules
through an LLM orchestrator without a dedicated deterministic rule layer;
(iii)~conflates model-class selection and invocation policy in the same tiered hierarchy;
(iv)~relies on LLM agents throughout---a limitation explicitly acknowledged by its
authors~\cite{kim2025tao}; (v)~has no metrological apparatus; and (vi)~does not address
model parsimony.
Eywa addresses the LLM-only limitation by formalising the FM--LLM interface but,
as a domain-collaboration framework, does not propose a layered trust architecture
or a metrologically grounded metric suite.

\textbf{Stream~3 -- Metrological uncertainty for ML.}
Bilson et al.~\cite{bilson2025metrology} transfer VIM/GUM apparatus to ML classification
via probability mass functions;
Thompson et al.~\cite{thompson2024trustworthy} bridge trustworthy AI and metrology
conceptually.
Both are limited to classification tasks without agentic architecture or cross-domain scope.

\textbf{Stream~4 -- Model parsimony.}
Belcak et al.\ (NVIDIA)~\cite{belcak2025slm} argue that small language models are the
appropriate medium for most agentic sub-tasks;
Schwartz et al.~\cite{schwartz2020greenai} frame efficiency as a first-class evaluation
metric.
These works provide an economic and engineering imperative but do not offer a trust
architecture.

Parallel governance anchors---NIST AI RMF~\cite{nist2023airmf},
NIST GAI Profile~\cite{nist2024gaiprofile},
ISO/IEC~42001~\cite{iso42001},
EU AI Act~\cite{euaiact2024}, and the cross-domain framework by
Herrera-Poyatos et al.~\cite{herrera2025responsible}---provide regulatory scaffolding
but not an engineering architecture.
TRACE positions itself as the engineering layer that operationalises those obligations
through measurable, architecture-embedded artefacts.

\textbf{Gap.}
No published work unifies (i)~a multi-layer trust architecture,
(ii)~cross-domain instantiation in operationally critical domains,
(iii)~a metrologically grounded trust-metric suite, and
(iv)~a model-parsimony principle in a single framework.
TRACE closes this four-sided gap.

\textbf{Genesis.}
TRACE emerged from the author's recognition and formalisation of a shared structural
pattern observed across two independently developed operationally critical systems:
(i)~a clinical decision-support architecture with metrological uncertainty
monitoring~\cite{zabolotnii2026clinical}, in which the author is a co-developer,
currently under review at \emph{IEEE Instrumentation \& Measurement Magazine};
and (ii)~an industrial multi-domain agentic-AI platform for upstream oil-and-gas
operations spanning technology/operations/administrative loops, developed independently
by Shcherban (2024--2026)~\cite{shcherban2025patent,shcherban2026usco}.
Recognising the four-layer decomposition, tiered escalation, and parsimony of learned
components shared between these two structurally distinct systems is the conceptual core
of TRACE; this paper formalises that recognition.

\textbf{Contributions.}
This work: (a)~presents TRACE as a domain-neutral engineering framework grounded in six
principles with explicit metrological analogues; (b)~defines a four-layer reference
architecture with an L2a/L2b split based on the parsimony principle and an
architecturally separate L3 orchestration-and-escalation policy; (c)~specifies a
layer-wise trust-metric suite culminating in CPR---the first quantitative first-class
parsimony metric in the trustworthy-AI literature; (d)~instantiates the framework in
three domains (clinical, industrial multi-domain, judicial) and makes the architectural
distinction from TAO precise.

% ─────────────────────────────────────────────────────────────────────────────
\section{The TRACE Framework}
\label{sec:trace}
% ─────────────────────────────────────────────────────────────────────────────

\subsection{Six Design Principles}
\label{sec:principles}

TRACE rests on six mutually reinforcing principles, each with a metrological analogue:

\begin{enumerate}

\item \textbf{Evidence Traceability.}
Every prescriptive action carries a machine-readable evidence trail.
\textit{Metrological analogue:} traceability of measurement results~\cite{jcgm200vim}.

\item \textbf{Bounded Human Supervision.}
Human oversight is an architectural layer with measurable workload and veto rights,
not a cosmetic safety net.
\textit{Analogue:} operator supervision of measurement systems~\cite{mosqueira2023hitl}.

\item \textbf{Staged Autonomy.}
Action rights are earned through accumulated evidence, not granted by default.
\textit{Analogue:} type approval and operational qualification.

\item \textbf{Bounded Context.}
The input context is explicitly specified, time-stamped, and kept current as a
safety property.
\textit{Analogue:} operational limits and environmental conditions.

\item \textbf{Metrological Accountability.}
All quality characteristics are specified, measurable, calibrated, and monitored
over time~\cite{bilson2025metrology,jcgm100gum}.
\textit{Analogue:} calibration, drift monitoring, stability testing.

\item \textbf{Model Parsimony / Right-Sized Components.}
The type of learned component (classical ML, specialised neural network, LLM, or
hybrid) is chosen by fitness for purpose, not by LLM presumption~%
\cite{belcak2025slm,schwartz2020greenai}.
\textit{Analogue:} \emph{fitness for purpose}---the foundational metrological
requirement.

\end{enumerate}

The acronym TRACE reflects five user-visible system properties
(\emph{trustworthy, reasoned, accountable, context-bound, escalated});
model parsimony operates as an internal design constraint.

\subsection{Four-Layer Reference Architecture}
\label{sec:arch}

\begin{figure}[t]
  \centering
  \includegraphics[width=0.95\linewidth]{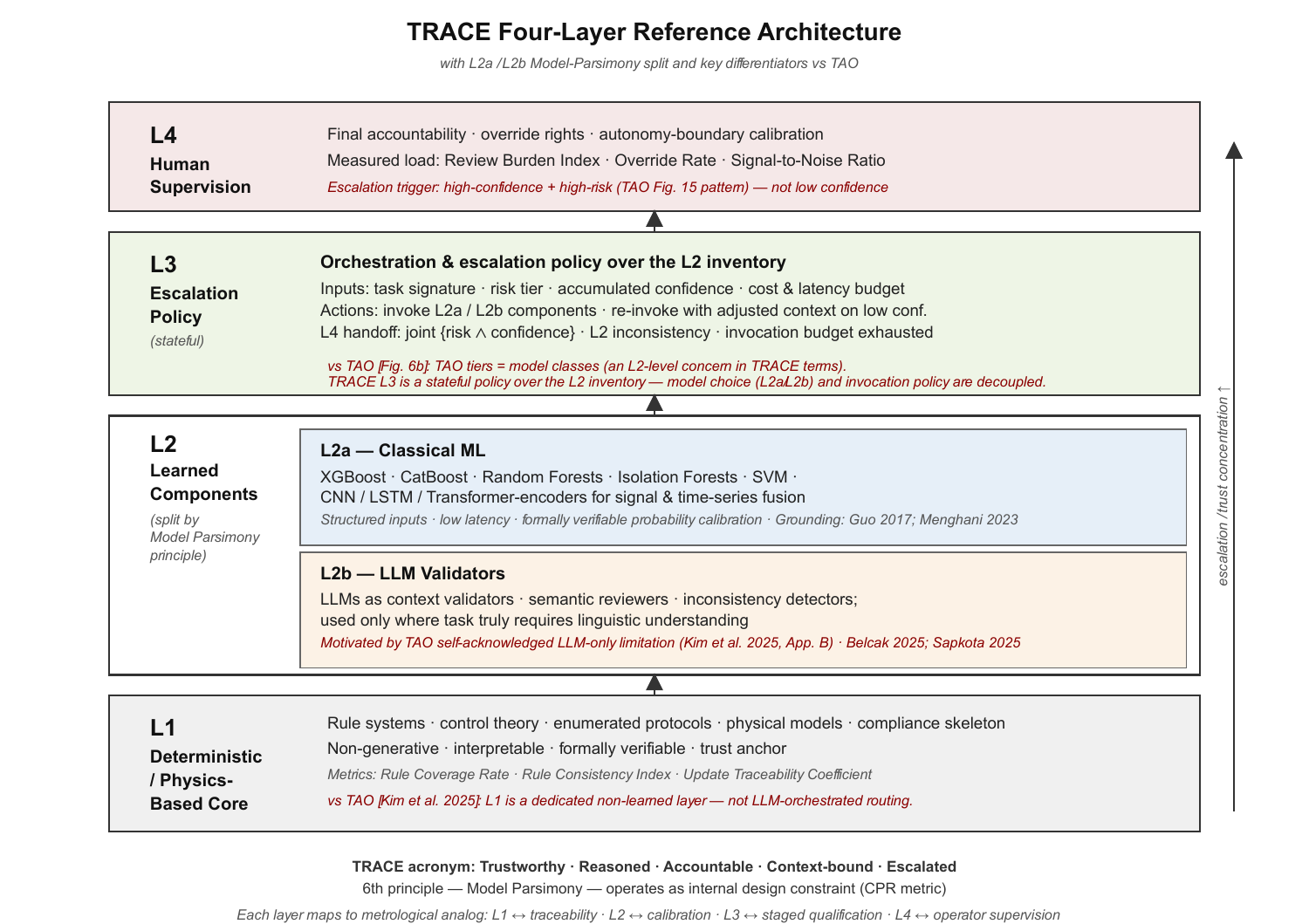}
  \caption{TRACE four-layer reference architecture.
    L1 provides the deterministic rule core (trust anchor);
    L2 holds the stateless learned-component inventory, partitioned into
    classical ML (L2a) and LLM validators (L2b);
    L3 is the stateful orchestration-and-escalation policy;
    L4 is bounded human supervision.}
  \label{fig:arch}
\end{figure}

Figure~\ref{fig:arch} shows the four-layer decomposition.

\textbf{L1 -- Deterministic / Physics-Based Core.}
Non-generative, interpretable, stable logic: rule systems, control theory, enumerated
protocols, physical models, compliance scaffolding.
\textit{Key architectural distinction from TAO~\cite{kim2025tao}}: TAO implements routing
and rules through an LLM orchestrator; TRACE separates L1 as an independent, non-learned,
formally verifiable layer and the primary trust anchor of the system.

\textbf{L2 -- Learned-Component Inventory (stateless).}
A catalogue of trained components partitioned by the parsimony principle.
L2 defines \emph{what} the system can do; \emph{when and in what order} components are
invoked is the responsibility of L3 policy.
Each L2 invocation is stateless and accountable to its calling policy.
\begin{itemize}
  \item \textbf{L2a -- Specialised / Classical ML.}
    Gradient boosting (XGBoost, CatBoost), random forests, isolation forests, SVM,
    and classical neural architectures (CNN, LSTM, Transformer encoders for sensor fusion).
    Applied where inputs are structured, latency is critical, reproducibility is
    required, and probability calibration is formally verifiable~%
    \cite{guo2017calibration,menghani2023efficient}.
  \item \textbf{L2b -- Generative / LLM Validators.}
    LLMs and generative models as contextual validators, semantic reviewers, and
    inconsistency detectors for free-form data.
    Invoked only where the task genuinely requires linguistic understanding.
\end{itemize}

The L2a/L2b boundary is described here at the architectural level. A concrete
protocol-level realisation of an analogous boundary is given by Eywa's
\emph{Tsaheylu} interface~\cite{li2026eywa}, which formalises the communication
contract between domain-specific foundation models (analogous to L2a) and
language-model agents (analogous to L2b). TRACE remains protocol-agnostic;
Tsaheylu is one viable implementation of the L2a$\leftrightarrow$L2b boundary
within the framework.

\textbf{L3 -- Orchestration-and-Escalation Policy (stateful).}
An architecturally separate policy layer \emph{above the L2 inventory}, operating on
task state (passage history, accumulated confidence, risk context, latency and cost
budget).
The policy specifies:
(i)~which L2 components are invoked and in what order for a given task type;
(ii)~conditions for re-invocation with expanded context or switching to an alternative
component;
(iii)~escalation triggers to L4---the \emph{joint} condition of high risk and high
confidence~\cite{kim2025tao} (a confident error in a high-risk context is more dangerous
than an acknowledged uncertainty), or L3-detected inconsistency between L2 components,
or exhaustion of the re-invocation budget;
(iv)~cost accounting for each execution path (Tier Cost Coefficient).

\textit{Key distinction from TAO}: TAO implements tiered oversight as a hierarchy of LLM
agents under an LLM orchestrator---in TRACE terms, this conflates L2-concern
(model-class selection) and L3-concern (invocation policy).
TRACE separates them: L2a vs.\ L2b is a static design decision by fitness for purpose;
invocation sequence, re-invocations, and escalation are L3 policy, independently
calibrated from operational statistics.
This separation removes the TAO safety-parsimony trade-off.

\textbf{L4 -- Bounded Human Supervision.}
Final accountability, veto, and calibration of AI authority bounds.
L4 entry is determined by L3 policy, not by direct L2 requests.

\subsection{Sub-Domain $\times$ Layer Matrix}
\label{sec:matrix}

Real deployments often span multiple operational sub-domains.
TRACE represents this as a \emph{sub-domain $\times$ layer matrix}: each sub-domain
instantiates all four layers independently, and cross-sub-domain coordination is provided
by a master orchestrator.
This yields systematic, transferable deployment templates for complex multi-domain
systems.
Instance~B (Section~\ref{sec:instanceB}) provides a documented precedent for this
matrix across three sub-domains with principally different data types.

\subsection{Trust Metric Suite}
\label{sec:metrics}

TRACE specifies 17 metrics: 12 layer-wise, 4 cross-cutting, and 1 parsimony metric.
The suite extends the metrological apparatus of
Bilson et al.~\cite{bilson2025metrology} and
Thompson et al.~\cite{thompson2024trustworthy} from ML classification alone to the full
agentic lifecycle:
\emph{ingestion $\to$ inference $\to$ escalation $\to$ human adjudication}.

\medskip
\noindent\textbf{Layer-wise metrics:}

\noindent\textit{L1:} Rule Coverage Rate, Rule Consistency Index,
Update Traceability Coefficient.

\noindent\textit{L2:} Context Relevance Precision, Context Freshness Index,
Input Perturbation Stability Rate.

\noindent\textit{L3:} Escalation Precision (correctness of L4 hand-off decisions);
Tier Cost Coefficient (aggregate computational cost of the policy-chosen execution path);
False Positive Attenuation (suppression of spurious escalations through L2
re-invocations).
These metrics evaluate the policy itself, not the quality of individual L2
components~\cite{dong2023reliability}.

\noindent\textit{L4:} Review Burden Index, Override Rate, Signal-to-Noise Ratio.

\medskip
\noindent\textbf{Cross-cutting metrics:}
Evidence Trail Completeness;
Calibration Error~\cite{guo2017calibration};
Autonomy Boundary Compliance;
Operational Stability Index.

\medskip
\noindent\textbf{Parsimony metric -- Computational Parsimony Ratio (CPR).}
CPR is the ratio of the resource cost of the most economical model that adequately
solves the sub-task (meeting required accuracy, calibration, and operational reliability)
to the resource cost of the deployed model.
$\text{CPR} = 1$ indicates an optimal model choice;
values substantially below~1 signal architectural overhead.
Resource cost encompasses latency, compute, training, and operational expenditure.
To the best of our knowledge, CPR is the first formalisation of Complexity-Performance
Ratio as a first-class design principle in the trustworthy-AI literature.
Empirical support for the underlying intuition is provided by Eywa's EywaBench
results~\cite{li2026eywa}: across nine scientific domains, replacing LLM-only
pipelines with FM--LLM heterogeneous collaboration improves utility while reducing
token cost---precisely the regime CPR is designed to surface as architectural overhead.

Each metric has a direct metrological analogue (accuracy, repeatability, systematic
error, drift, fitness for purpose), enabling principled aggregation via GUM-style
uncertainty propagation across layers~\cite{jcgm100gum,bilson2025metrology}.

% ─────────────────────────────────────────────────────────────────────────────
\section{Multi-Domain Instantiations}
\label{sec:instances}
% ─────────────────────────────────────────────────────────────────────────────

The framework's universality is demonstrated by three instantiations.
Instances A and B are \emph{foundational precedents}---independently developed
systems whose shared structural pattern motivated TRACE's formalisation.
Instance~C is a \emph{partially implemented extension} demonstrating architectural portability.

\subsection{Instance A: Clinical Decision Support (Foundational)}
\label{sec:instanceA}

A detailed account of the clinical foundational implementation---architecture, metrics,
validation---appears in a companion paper~\cite{zabolotnii2026clinical} currently under
review at \emph{IEEE Instrumentation \& Measurement Magazine}.
The TRACE layer projection is:
\begin{itemize}
  \item \textbf{L1:} Rule-based clinical logic, compatible with the FUTURE-AI
    traceability principle~\cite{lekadir2025futureai}.
  \item \textbf{L2a:} Classical ML for structured prediction: risk scores,
    vital-sign time-series classification, laboratory data modelling.
  \item \textbf{L2b:} LLM validator for free-form clinical notes;
    contextual consistency checking of recommendations against patient history.
  \item \textbf{L3:} Clinical escalation policy: routine cases are handled by L2a
    risk scores with calibrated probabilities; at boundary confidence or with a
    complex context, L2b is invoked for free-form data verification; on joint
    high-risk / high-confidence coincidence, or L3-detected inconsistency between
    L2a and L2b, mandatory clinician hand-off is triggered.
    Compatible with the TAO pattern~\cite{kim2025tao} but with explicit L1/L2/L3
    separation.
  \item \textbf{L4:} Clinician as final reviewer, aligned with the TRIAD
    collaboration framework~\cite{li2026triad}.
\end{itemize}

\subsection{Instance B: Industrial Multi-Domain Platform (Foundational)}
\label{sec:instanceB}

\begin{figure}[t]
  \centering
  \includegraphics[width=0.95\linewidth]{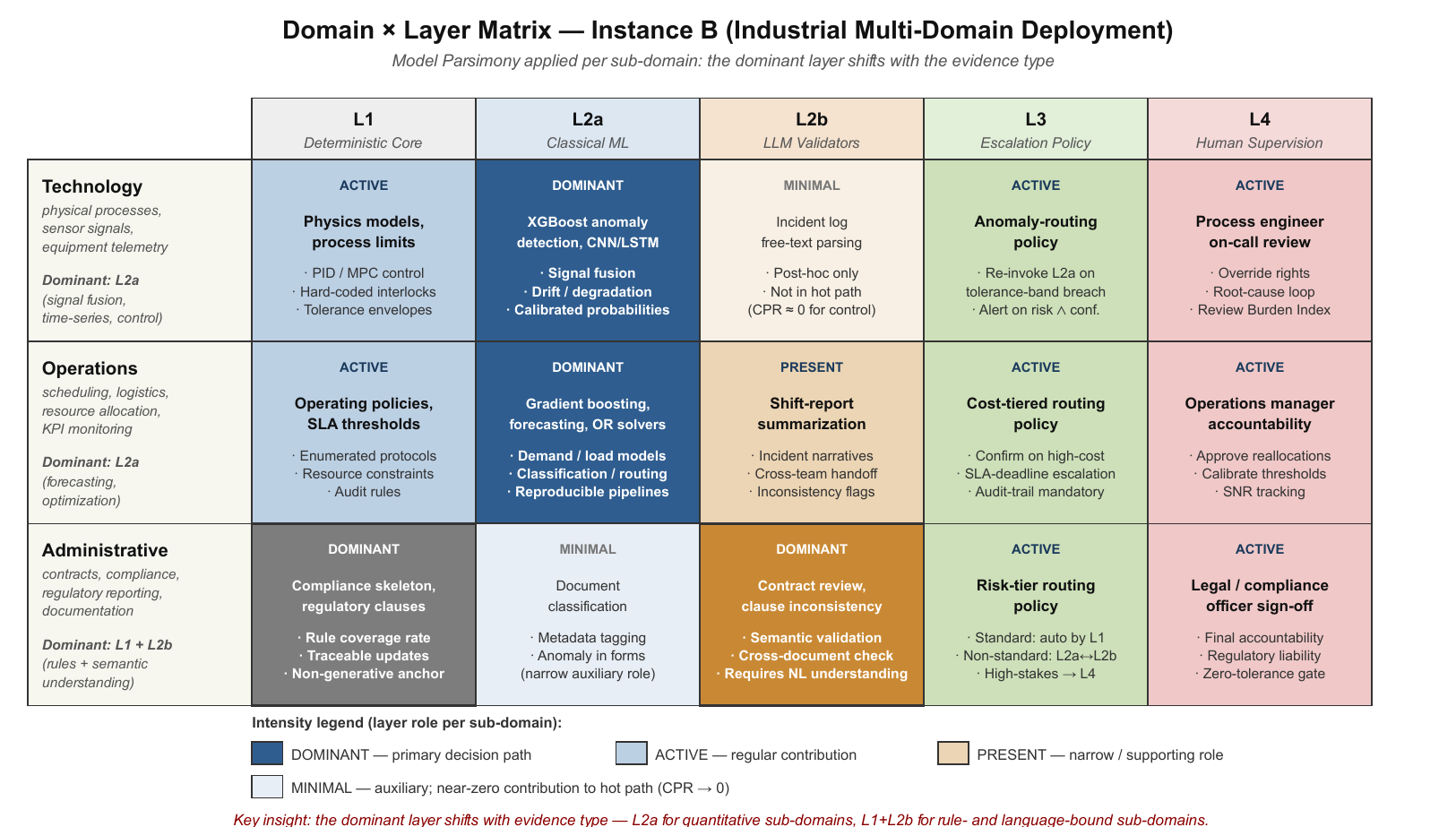}
  \caption{Sub-domain $\times$ layer matrix for Instance~B (industrial multi-domain
    platform). The same four-layer architecture manifests differently across three
    sub-domains: in Technology and Operations, L2a (classical ML) dominates;
    in Administrative, the dominant pair is L1 (compliance rules) and L2b
    (semantic contract review). Model parsimony applies adaptively to the evidence
    type of each sub-domain. L3 is realised as an independent Escalation Policy
    in each sub-domain; a master orchestrator coordinates cross-sub-domain pipelines.}
  \label{fig:matrix}
\end{figure}

\textbf{Attribution and scope.}
Instance~B is an independently developed industrial agentic-AI platform for upstream
oil-and-gas operations, designed by Shcherban
(2024--2026)~\cite{shcherban2025patent,shcherban2026usco}.
The system simultaneously spans three operational sub-domains:
\textbf{technology} (drilling, production, well operations),
\textbf{operations} (maintenance, condition monitoring), and
\textbf{administrative} (procurement, contract lifecycle, compliance).
This system serves here as a structural reference exhibiting the TRACE pattern in a
domain principally different from clinical decision support; detailed methodology,
operational statistics, and component-level specifications are reported separately by
the system's developer (Paper~2, in preparation).

\textbf{Architectural projection on TRACE layers.}
At the level of architectural pattern visible in the public record of the system, each
sub-domain instantiates all four TRACE layers, but with characteristically different L2
weighting (Figure~\ref{fig:matrix}):
the \emph{technology} and \emph{operations} sub-domains are \textbf{L2a-dominant}, since
the bulk of inference operates over structured, low-latency sensor and equipment-state
signals where classical and specialised ML satisfy calibration and reproducibility
requirements at substantially lower cost than LLMs;
the \emph{administrative} sub-domain is \textbf{L1+L2b-dominant}, since regulatory and
contractual rules must be formally specified rather than learned (L1), while semantic
review of free-form contract text and unstructured documents is the natural target for
LLM validators (L2b).
A master orchestrator coordinates cross-sub-domain pipelines
(e.g., the chain procurement~$\to$~contract~$\to$~compliance check~$\to$~operational
integration when new downhole equipment is introduced) and aggregates TRACE metrics
across sub-domains, enabling measurement of operational stability for the platform as a
whole.

\textbf{Significance for TRACE.}
At the level of structural pattern, Instance~B demonstrates three claims that motivated
TRACE's formalisation:
(i)~the parsimony principle is empirically natural---two of three sub-domains are
L2a-dominant despite the broader trend of LLM-default deployment;
(ii)~the sub-domain $\times$ layer matrix (\S\ref{sec:matrix}) is documented in a
real-world design precedent;
(iii)~the architecture transfers across principally different evidence types (sensor
signals vs.\ legal documents) within a single platform.
Component-level validation of these claims belongs to the domain paper.

\subsection{Instance C: Judicial AI Assistant (Partially Implemented)}
\label{sec:instanceC}

L1---procedural and substantive legal norms~\cite{cepej2018charter,%
cepej2023operationalisation}.
L2a---classical classifiers for document categorisation, precedent relevance ranking,
NER for parties and circumstances.
L2b---LLM for case-material analysis, precedent semantic correlation, draft document
preparation.
Empirical anchor: Magesh et al.~\cite{magesh2025hallucination} report a 17\%
hallucination rate in commercial legal RAG systems, providing direct motivation for
multi-layer trust oversight.
L3---legal escalation policy: L2a/NER retrieval selects relevant precedents;
L2b mandatory hallucination verification; at high claim stakes and a confident
recommendation, juridical L4 review is required before execution.
L4---legal professional (judge, assistant)~\cite{kolkman2024justitia}.

Instance~C is currently partially implemented as an AI assistant for judicial-decision
analysis within the EU project ``Pravo-Justice'' --- ``Support of the Supreme
Court on improving (modernising) the Database of Legal Positions''
(funded by Expertise France)~\cite{pravojustice3,pravojustice2024db}.
This serves as a demonstration of architectural portability: the same four-layer pattern
and the same metrics apply in a domain with a principally different governance context
(judicial ethics, burden of proof, procedural accountability).

\subsection{Unity and Variation}
\label{sec:unity}

Across all three instances the \emph{same reference architecture, the same six principles,
and the same metric suite} apply---only the instantiated parameters and the L2a/L2b
split vary.
Instance~B additionally demonstrates that the sub-domain $\times$ layer matrix operates
within a single system for three sub-domains with principally different data types.
This invariance is the central empirical claim of the framework.

% ─────────────────────────────────────────────────────────────────────────────
\section{Discussion}
\label{sec:discussion}
% ─────────────────────────────────────────────────────────────────────────────

\subsection{Relation to TAO, Revisited}
\label{sec:tao}

TAO~\cite{kim2025tao} supplies two empirical facts that TRACE preserves and absorbs:
(i)~tiered oversight reduces individual-agent error rates by 16.9--24.3\% (Table~6);
(ii)~confident errors in high-risk contexts are more dangerous than acknowledged
uncertainty (Fig.~15).
TRACE reinterprets both within a purely layered architecture:
(i)~becomes the lower bound on the expected effect of any correctly configured L3 policy
operating over a heterogeneous L2 inventory;
(ii)~becomes the precise specification of the L3$\to$L4 hand-off trigger
(joint risk $\wedge$ confidence).

TAO's descending-vs-ascending capability finding---that descending is safer---is
reinterpreted as valid \emph{within a single model class} and serves as a calibration
anchor for L3 policy, not as an architectural norm.
This interpretation is only visible after separating L2-concern (model-class selection)
from L3-concern (invocation policy), which TAO does not do.

\subsection{Relation to Eywa: Complementary Decomposition}
\label{sec:eywa}

Eywa~\cite{li2026eywa} and TRACE address overlapping problems from complementary
angles. Eywa provides a \emph{protocol-level} answer to the question of how
heterogeneous foundation models and LLMs should communicate, with three concrete
instantiations (EywaAgent, EywaMAS, EywaOrchestra) and an empirical benchmark
(EywaBench) over nine scientific domains. TRACE provides an \emph{architectural and
metrological} answer to the question of what layer-wise decomposition operationally
critical AI should adopt, and how the resulting trust properties should be measured.
The two are not competing: Eywa's Tsaheylu protocol is a viable realisation of the
TRACE L2a$\leftrightarrow$L2b boundary; conversely, TRACE supplies the L1
deterministic anchor, the L3 stateful policy, the L4 supervision layer, and the
GUM/VIM-grounded metric suite that Eywa, by design, does not address.
A natural integration path---layered TRACE architecture with a Tsaheylu-conformant
L2a/L2b interface---is left to follow-up work.

\subsection{Parsimony as an Architectural Constraint}
\label{sec:parsimony-disc}

CPR introduces a quantitative parsimony metric into trustworthy-AI alongside accuracy,
calibration, and review burden.
Its practical value is twofold: the metric surfaces architectural overhead (an LLM
deployed where a gradient-boosted classifier satisfies the required calibration yields
CPR $\ll 1$) and provides an auditable artefact for certification bodies
(e.g., under EU AI Act Art.~15~\cite{euaiact2024}).
The architectural correlate of CPR is the L2a/L2b split: it makes the parsimony choice
a \emph{visible} design decision rather than an implicit LLM-default bias.
Instance~B makes this concrete: two of three sub-domains are naturally L2a-dominant;
an LLM-by-default deployment would be measurably over-engineered.

\textbf{Limitations.}
The Trust Metric Suite is currently specified analytically; empirical calibration from
production deployments is forthcoming in the domain-specific follow-up papers.
CPR requires operational benchmarking for normalisation against domain-specific cost
models.
Instance~C is partially implemented; the EU project deployment covers judicial-decision
analysis but full operational statistics await the completion of Phase~2.

% ─────────────────────────────────────────────────────────────────────────────
\section{Conclusion}
\label{sec:conclusion}
% ─────────────────────────────────────────────────────────────────────────────

TRACE provides a metrologically grounded, cross-domain engineering framework for
trustworthy agentic AI in operationally critical domains.
Its four-layer architecture, six design principles, 17-metric trust suite, and CPR
parsimony metric synthesise four previously disconnected literature streams into a single
reusable engineering blueprint.
Two foundational precedents---clinical decision support and an industrial
multi-domain platform---validate the shared structural pattern that motivated
TRACE's formalisation.

TRACE operationalises trust and model parsimony as engineering outcomes, shifting the
design question from ``how intelligent is this model?'' to ``what is this system
permitted to do, on what grounds, under what supervision, with what measurable
uncertainty, and is the right type of model being used for each sub-task?''

% ─────────────────────────────────────────────────────────────────────────────
\section*{Acknowledgements}
% ─────────────────────────────────────────────────────────────────────────────

The author thanks \textbf{Andriy Shcherban}, whose independently developed industrial
multi-domain platform served as the empirical catalyst for TRACE's formalisation;
the structural pattern shared between that system and the clinical foundational
implementation motivated the abstraction presented here.

The clinical foundational implementation (Instance~A) is reported separately by
S.~Zabolotnii, V.~Holinko, and O.~Antonenko~\cite{zabolotnii2026clinical};
the author thanks V.~Holinko and O.~Antonenko for the collaboration on that companion
work.

% ─────────────────────────────────────────────────────────────────────────────
\bibliographystyle{unsrtnat}
\bibliography{references_TRACE}
% ─────────────────────────────────────────────────────────────────────────────

\end{document}